\documentclass[11pt,a4paper]{article}
\PassOptionsToPackage{usenames, dvipsnames, table}{xcolor}
\usepackage[hyperref]{acl2021}
\usepackage{times}
\usepackage{latexsym}
\usepackage{booktabs, multirow} 
\usepackage{soul}
\usepackage{adjustbox}
\usepackage{changepage,threeparttable} 
\usepackage{tabularx}

\usepackage{microtype}

\usepackage[capitalize]{cleveref}
\usepackage{float}
\usepackage{hyperref}

\usepackage[markup=bfit, deletedmarkup=sout, authormarkup=superscript]{changes}
\definechangesauthor[name={td}, color=red]{TODO}
\definechangesauthor[name={ar}, color=ForestGreen]{AR}

\aclfinalcopy 
\def\aclpaperid{3675} 

\title{PROST:\\
Physical Reasoning about Objects through Space and Time}

\author{Stéphane Aroca-Ouellette\thanks{*Email has no accent, but includes the hyphen} , Cory Paik, Alessandro Roncone \and Katharina Kann \\
  University of Colorado Boulder\\
  \texttt{firstname.lastname@colorado.edu}}
  
\date{May 31st 2021}

\begin{document}
\maketitle
\begin{abstract}

We present a new probing dataset named \emph{PROST: Physical Reasoning about Objects Through Space and Time}. 
This dataset contains 18,736 multiple-choice questions made from 14 manually curated templates, covering 10 physical reasoning concepts. All questions are designed to probe both causal and masked language models in a zero-shot setting. We conduct an extensive analysis which demonstrates that state-of-the-art pretrained models are inadequate at physical reasoning: they are influenced by the order in which answer options are presented to them, they struggle when the superlative in a question is inverted (e.g., \textit{most} $\leftrightarrow$ \textit{least}), and increasing the amount of pretraining data and parameters only yields minimal improvements. These results provide support for the hypothesis that current pretrained models' ability to reason about physical interactions is inherently limited by a lack of real world experience. By highlighting these limitations, we hope to motivate the development of models with a human-like understanding of the physical world.

\end{abstract}

\section{Introduction}
In the context of natural language processing (NLP), \citet{climbing} provides a working definition of ``understanding" as the ability to recover the communicative intent from an utterance. To achieve this, one must be able to query a set of concepts that is aligned with the speaker's own understanding. An example of such alignment is our interaction with the physical world. This experience, shared by all humans, provides a common set of concepts to rely on in communication.
For example, the reader can map the phrase \emph{I dropped my pint glass} to a set of relevant experiences and generate a mental depiction of the scene. Further, the reader can also use their knowledge of gravity and the properties of a pint glass to reason about potential outcomes: the pint glass will fall toward the ground and will likely break on impact. 
\begin{figure}[t!]
    \centering
    \includegraphics[width=\columnwidth]{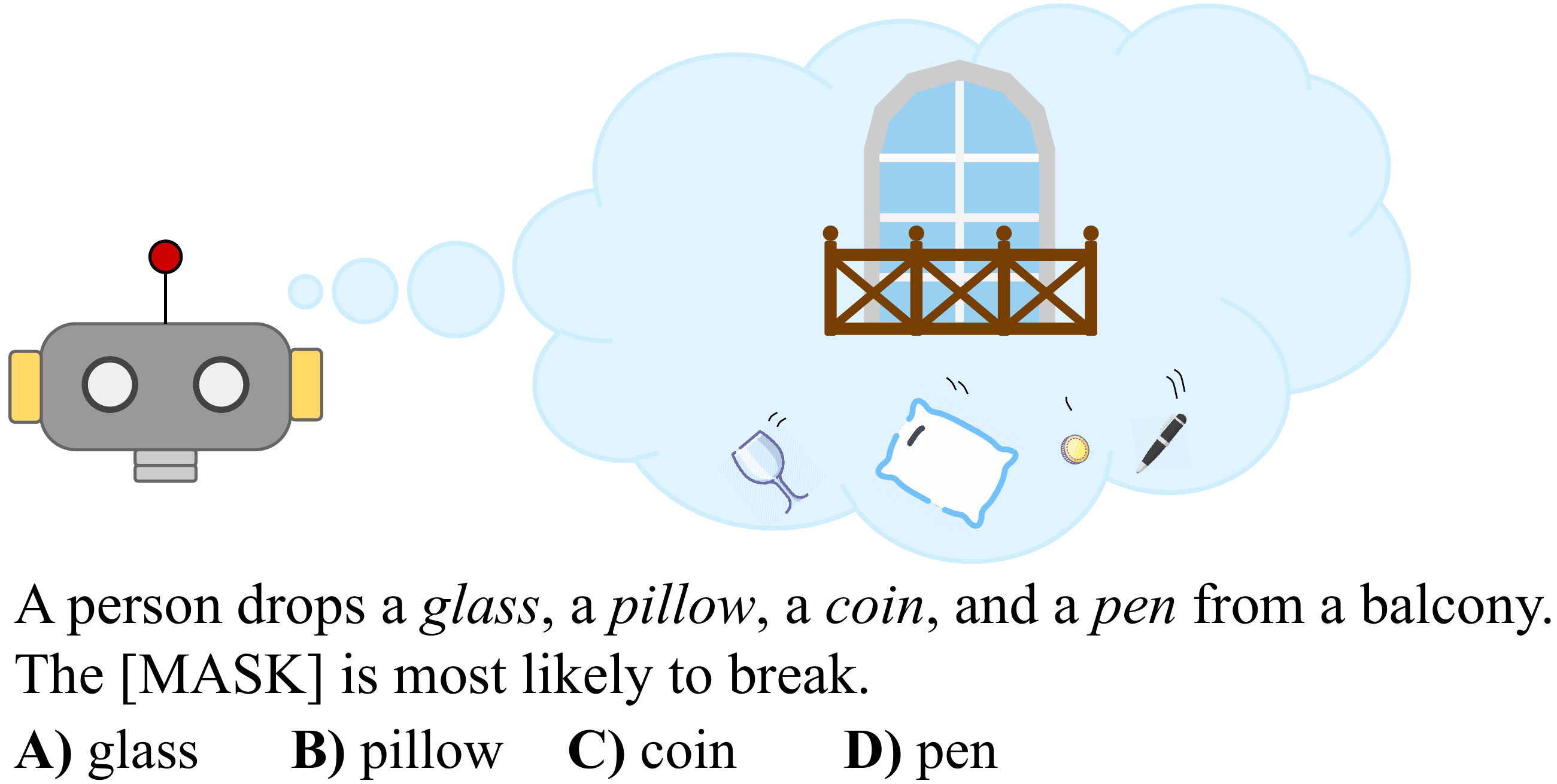}
    \caption{An example question from PROST.}
    \label{fig:example1}
\end{figure}

Children grab, push, and play with the objects around them to form concepts about the world they live in even before learning to talk \citep{precursor}. These concepts are then linked with words to enable communication, eventually providing the necessary grounds for concepts and language to co-develop \cite{learn_words, concepts}. 
In contrast, current language models (LMs) are not exposed to real-world experiences, making them incapable of grounding language \cite{experience}. We hypothesize that this lack of experience impedes their ability to both understand an utterance relating to the physical world and their ability to reason about its implications.

In order to investigate our hypothesis, we create \emph{PROST: Physical Reasoning of Objects Through Space and Time}, a probing dataset to evaluate the ability of pretrained LMs to understand and reason about the physical world.
PROST consists of multiple-choice cloze-style questions covering $10$ basic concepts: direction, mass, height, circumference, stackable, rollable, graspable, breakable, slideable, and bounceable.
Importantly, PROST is designed to avoid models succeeding in unintended ways. First, PROST provides no training data, so as to probe models in a zero-shot fashion. This prevents models from succeeding through spurious correlations between training and test data and encourages success through a true understanding of and reasoning about the concepts at hand. Second, we manually write templates for all questions in an effort to prevent models from having seen the exact same sentences in their training data. Finally, it focuses on a small set of well defined, objective concepts that only require a small vocabulary. This allows researchers to focus more on the quality of training data rather than the size of it.

\paragraph{Contributions} We make two contributions: 1) We introduce PROST, a dataset with $18,736$ cloze-style questions created from $14$ manually written templates, covering $10$ physical reasoning tasks.
2) We conduct an extensive analysis which demonstrates that state-of-the-art pretrained models are inadequate at physical reasoning.
More specifically, they are influenced by the order in which answer options are presented to them, they struggle when the superlative in a question is inverted (e.g., \textit{most} $\leftrightarrow$ \textit{least}), and increasing the amount of pretraining data and parameters only yields minimal improvements. The dataset and code is available at \href{https://github.com/nala-cub/prost}{github.com/nala-cub/prost}. 

\section{Related Work}
\paragraph{Evaluation of Reasoning Abilities}
As pretrained models are excelling on many NLP tasks, more work is being done on understanding their abilities.
A subset of this work focuses on physical reasoning. 
PIQA \citep{PIQA} tests physical commonsense, with
concepts ranging from hard shell tacos to separating egg yolks. 
In order to succeed on PIQA through reasoning, a model would need to be able to understand thousands of human experiences. In contrast, PROST provides a first step towards grounded understanding and reasoning by focusing on a few simple concepts. \citet{PHYRE} provides a set of 2D puzzles that involve placing a new object in a scene to accomplish a goal. This research also focuses on simple physics, however there is no language component. 
\citet{ARC} and \citet{TQA} both provide a large set of grade school multiple-choice questions, including some that could be solved with reasoning. However both provide corresponding material where the solution can be found, relying more on information retrieval than a general understanding and reasoning about the world.

Another set of reasoning-based benchmarks focuses on common sense reasoning. SWAG and its extension hellaSWAG evaluate commonsense natural language inference \cite{swag,hellaswag}. 
\citet{socialIQA} tests commonsense reasoning about social situations.
However, commonsense reasoning is often subjective and requires understanding of complex human--human interactions involving social and societal norms. In contrast, physical reasoning is based on objective and well defined constructs.
Other datasets \citep{verb_phys, lions, goel_words} focus on object--attribute comparison. However, they compare concepts at a word level rather than sentence level and use a large training set to create an engineered object--attribute comparison model. It is difficult to see how these models could generalize to other forms of reasoning.

Moreover, all the above datasets follow a pretraining-agnostic identically distributed (PAID) paradigm \cite{PAID}, making them susceptible to models that can leverage unintended correlations between the training and test sets.
\paragraph{Zero-Shot LM Probes}
Similar to PROST, several recent benchmarks have circumvented the concern of identically distributed training and test sets by probing models in a zero-shot manner. \citet{LAMA} queries masked LMs (MLMs) for factual knowledge using templates in the format of \emph{Dante was born in [MASK].} \citet{olmpics} use a similar format
to probe six concepts ranging from age comparison to taxonomy conjunction. \citet{BERTisNot} uses this format to show that BERT robustly retrieves hypernyms, but fails to understand negation. \citet{NumerSense} probe numerical commensense in both MLMs and traditional LMs. 
\citet{blimp}
measures traditional LMs' sense of grammatical acceptability by comparing sentence probabilities. 

\paragraph{Grounded Language Environments}
PROST investigates if pretrained models show a lack of understanding of the physical world which could result from learning language without grounding. While not used for pretraining,
a number of multi-modal environments have been developed to ground language. \citet{ALFRED}'s ALFRED builds on other vision-and-language navigation environments \citep{IQA, TaCOS, VSP, R2R}, and enables grounding of language instruction to actions, behaviours, and objects. BABYAI \cite{BABYAI} and BABYAI++ \citep{BABYAI++} provide an environment to ground simple language in a gridworld.
Additionally, other work has explored grounding language in simulations or the real world \citep{grounded_fast_slow, grounding_play}. While they provide important resources to ground language, little emphasis is placed on the language modules themselves. They are often trained tabulae rasae, learning language for a singular purpose and missing out on the syntax and coverage learnt during pretraining;\footnote{An exception is \citet{grounding_play}, which incorporates modern LMs and provides impressive generalizability. However, they too only use language as an input and do not analyze how language understanding evolves.} language is only ever an input, 
and no analysis has been done on how language understanding evolves as the agent learns to succeed on different tasks.

\begin{table*}\footnotesize
\setlength{\tabcolsep}{3.5pt}
\centering
\begin{tabularx}{\textwidth}{l|c|c X}
\toprule
\textbf{Category} & \textbf{Qs.} & \multicolumn{2}{c}{\textbf{Template}}  \\
\midrule
Directs. 1 & 12 & \textbf{C:} &
A person is walking \textit{\{north/east/south/west\}}. They turn \textit{\{left/right/around\}}. \\
 &  & \textbf{Q:} &
They are now walking [MASK].  \\
& & \textbf{O:} &
\textbf{A)} north \textbf{B)} east \textbf{C)} south \textbf{D)} west\\

Directs. 2a & 1 & \textbf{C:} & A person drops a ball. \\
            &   & \textbf{Q:} & Immediately after leaving the person's hand, the ball is moving toward the [MASK]. \\

Directs. 2b & 1 & \textbf{C:} & A person throws a ball straight into the air. \\
            &   & \textbf{Q:} & Immediately after leaving the person's hand, the ball is moving toward the [MASK]. \\

Directs. 2c & 1 & \textbf{C:} & A person throws a ball straight into the air. \\
            &   & \textbf{Q:} & Immediately after reaching the highest point in it's trajectory, the ball is moving toward the [MASK]. \\
Directs. 2d & 1 & \textbf{C:} & A person drops a ball. The ball then bounces off the ground. \\
            &   & \textbf{Q:} & Immediately after bouncing off the ground, the ball is moving toward the [MASK]. \\
            &   & \textbf{O:} & \textbf{A)} ground \textbf{B)} sky \textbf{C)} left \textbf{D)} right\\
\midrule
Mass 1 & 720 & \textbf{C:} & A(n) \textit{\{mass\_obj1\}}, a(n) \textit{\{mass\_obj2\}}, a(n) \textit{\{mass\_obj3\}}, and a(n) \textit{\{mass\_obj4\}} moving at identical speeds each collide with a static hockey puck. \\
       &     & \textbf{Q:} & The puck hit by the [MASK] slides the \textit{\{shortest/longest\}} distance. \\

Mass 2 & 720 & \textbf{C:} & A(n) \textit{\{mass\_obj1\}} and a(n) \textit{\{mass\_obj2\}} are placed on either end of a perfectly balanced seesaw. \\
& & \textbf{Q:} &  The side of the seesaw  with the [MASK] moves \textit{\{up/down\}}. \\
& & \textbf{O:} & \textbf{A)} \textit{\{mass\_obj1\}} \textbf{B)} \textit{\{mass\_obj2\}} \textbf{C)} \textit{\{mass\_obj3\}} \textbf{D)} \textit{\{mass\_obj4\}}\\
\midrule
Height 1 & 720 & \textbf{C:} & 
Four balls are dropped. The first is dropped from the height equivalent of a \textit{\{height\_obj1\}}, the second is dropped from the height equivalent of a \textit{\{height\_obj2\}}, the third is dropped from the height equivalent of a \textit{\{height\_obj3\}}, and the fourth is dropped from the height equivalent of a \textit{\{height\_obj4\}}. \\
& & \textbf{Q:} & The ball dropped from the height of the [MASK] takes the \textit{\{longest/shortest\}} amount of time to fall. \\

Height 2 & 720 & \textbf{C:} &  There are four staircases. The first staircase leads to the top of a \textit{\{height\_obj1.\}}, the second staircase leads to the top of a \textit{\{height\_obj2.\}}, the third staircase leads to the top of a \textit{\{height\_obj3.\}}, and the fourth staircase leads to the top of a \textit{\{height\_obj4.\}}. \\
&&\textbf{Q:}& The staircase leading to the top of the [MASK] is the {easiest/hardest} to walk up. \\
&&\textbf{O:}& \textbf{A)} \textit{\{height\_obj1\}} \textbf{B)} \textit{\{height\_obj2\}} \textbf{C)} \textit{\{height\_obj3\}} \textbf{D)} \textit{\{height\_obj4\}}\\

\midrule
Circumf. 1 & 720 & \textbf{C:} & Four people are walking at identical speeds. The first walks around a \textit{\{circ\_obj1\}}, the second walks around a \textit{\{circ\_obj2\}}, the third walks around a \textit{\{circ\_obj3\}}, and the fourth walks around a \textit{\{circ\_obj4\}}. \\
&& \textbf{Q:} & The [MASK] takes the \textit{\{longest/shortest\}} amount of time to walk around. \\

Circumf. 2 & 720 & \textbf{C:} & 
 A person paints a circle around a \textit{\{circ\_obj1\}},  a \textit{\{circ\_obj1\}},  a \textit{\{circ\_obj1\}}, and a \textit{\{circ\_obj1\}}. \\
&&\textbf{Q:}  & The circle around the [MASK] takes the \textit{\{most/least\}} amount of paint. \\
&&\textbf{O:}  &  \textbf{A)} \textit{\{circ\_obj1\}} \textbf{B)} \textit{\{circ\_obj2\}} \textbf{C)} \textit{\{circ\_obj3\}} \textbf{D)} \textit{\{circ\_obj4\}}\\

\midrule
Stackable & 2400 & \textbf{C:} & 
A person is trying to stack \textit{\{stack\}}, \textit{\{no stack1\}}, \textit{\{no stack2\}}, and \textit{\{no stack3\}}. \\
&& \textbf{Q:}& The [MASK] are the \textit{\{easiest/hardest\}} to stack. \\
&& \textbf{O:}& \textbf{A)} \textit{\{stack\}} \textbf{B)} \textit{\{no stack1\}} \textbf{C)} \textit{\{no stack2\}} \textbf{D)} \textit{\{no stack3\}} \\

\midrule
Rollable & 2400 & \textbf{C:} &
A person is trying to roll a(n) \textit{\{roll\}}, a(n) \textit{\{no roll1\}}, a(n) \textit{\{no roll2\}}, and a(n) \textit{\{no roll3\}}. \\
&& \textbf{Q:} & The [MASK] is the \textit{\{easiest/hardest\}} to roll. \\
&& \textbf{O:} &  \textbf{A)} \textit{\{roll\}} \textbf{B)} \textit{\{no roll1\}} \textbf{C)} \textit{\{no roll2\}} \textbf{D)} \textit{\{no roll3\}} \\

\midrule
Graspable & 2400 & \textbf{C:} &
A person is trying to move a pile of \textit{\{break\}}, a pile of \textit{\{no break1\}}, a pile of \textit{\{no break2\}}, and a pile of \textit{\{no break3\}} from one side of a room to the other using only one hand. \\
&& \textbf{Q:} & The [MASK] is the \textit{\{most/least\}} likely to break.  \\
&& \textbf{O:} & \textbf{A)} \textit{\{break\}} \textbf{B)} \textit{\{no break1\}} \textbf{C)} \textit{\{no break2\}} \textbf{D)} \textit{\{no break3\}} \\

\midrule
Breakable & 2400 & \textbf{C:} & A person drops a \textit{\{break\}}, a \textit{\{no break1\}}, a \textit{\{no break2\}}, and a \textit{\{no break3\}} from a balcony. \\
&& \textbf{Q:} & The [MASK] is the \textit{\{most/least\}} likely to break.  \\
&& \textbf{O:} & \textbf{A)} \textit{\{grasp\}} \textbf{B)} \textit{\{no grasp1\}} \textbf{C)} \textit{\{no grasp2\}} \textbf{D)} \textit{\{no grasp3\}} \\

\midrule
Slideable & 2400 & \textbf{C:} &
A person is sliding four bricks across four hard surfaces. The first surface is covered with \textit{\{slide\}}, the second surface is covered with \textit{\{no slide1\}}, the third surface is covered with \textit{\{no slide2\}}, and the fourth surface is covered with \textit{\{no slide3\}}. \\
&& \textbf{Q:} & The surface covered with [MASK] is the \textit{\{hardest/easiest\}} for the brick to slide across. \\
&& \textbf{O:} & \textbf{A)} \textit{\{slide\}} \textbf{B)} \textit{\{no slide1\}} \textbf{C)} \textit{\{no slide2\}} \textbf{D)} \textit{\{no slide3\}}  \\

\midrule
Bounceable & 2400 & \textbf{C:} &
A person is trying to bounce a rubber ball. They drop a first ball onto \textit{\{bounce\}}, a second ball onto \textit{\{no bounce1\}}, a third ball onto \textit{\{no bounce2\}}, and a fourth ball onto \textit{\{no bounce3\}}. \\
&& \textbf{Q:} & The ball dropped onto[MASK] bounces the \textit{\{most/fewest\}} times.  \\
&& \textbf{O:} & \textbf{A)} \textit{\{bounce\}} \textbf{B)} \textit{\{no bounce1\}} \textbf{C)} \textit{\{no bounce2\}} \textbf{D)} \textit{\{no bounce3\}}  \\

\bottomrule
\end{tabularx}

\caption{All templates in PROST. \textbf{C:} = Context, \textbf{Q:} = Question, \textbf{O:} = Options. \{\} indicate placeholders. The objects can be found in \cref{tab:aff_objs}. The rest of the placeholders show their possibilities in the braces themselves.  [MASK] indicates the position of the blank that the models need to fill. See \cref{sec:templates} for more information. \newline \textit{NOTE: The number of objects with and without the affordances are swapped  when the superlative is inverted.}}
\label{tab:templates}
\end{table*}

\newcommand*{\escape}[1]{\texttt{\textbackslash#1}}
\begin{table*}[th!]\footnotesize
\centering
\begin{tabular}{l|l|l}
\toprule
Model Type & Input & Target \\
\midrule
Decoder & \textit{(context)} They are now walking $\mathbf{\left<O\right>}$.  & Max probability for each sentence input to the model  \\
\midrule
Encoder &\textit{(context)} They are now walking $\mathbf{\left<M\right>}$. & Max probability for the masked token.\\
\midrule
T5 & \textit{(context)} They are now walking $\mathbf{\left<X\right>}$. & $\mathbf{\left<X\right>}$ south $\mathbf{\left<Y\right>}$ \\
\midrule
\multirow{ 2}{*}{UnifiedQA} & Which way are they walking now? \escape{n} (A) north & south \\
& (B) south (C) east (D) west \escape{n} \textit{(context)}  \\
\bottomrule
\end{tabular}
\caption{Overview of the task preprocessing for different architectures evaluated. In all methods, the context remains unchanged and is ``A person is walking west. They turn left.''}
\label{tab:processing}
\end{table*}

\section{PROST}
\label{sec:templates}
PROST consists of $18,736$ cloze-style multiple-choice questions designed for probing a LM's physical reasoning ability.
They cover 10 basic concepts: direction, mass, height, circumference, stackable, rollable, graspable, breakable, slideable, and bounceable. We choose these concepts because they are well defined, easily learned by interacting with the world, and are useful concepts for any embodied agent. The questions are constructed from $14$ manually written templates. Each template follows one of three different formats: the first format is specific to the set of questions pertaining to directions; the second format is used to gauge the relative attributes---specifically mass, height, and circumference---of objects; and the third format targets the affordances of objects---specifically whether an object is stackable, rollable, graspable, or breakable, and whether a surfaces is slideable or bounceable\footnote{\emph{Bounceable} here refers to providing an elastic collision.}.
We use CheckList \citep{checklist} to obtain the questions from our templates. 
We show all templates in \cref{tab:templates} and explain them in detail below. We end this section by describing the objects featured in PROST.
\paragraph{Direction Templates}
We use two templates to generate questions which probe understanding of direction. The first focuses on cardinal directions. The second uses a set of four manually crafted questions to probe understanding of how gravity affects the directions of a ball throughout its trajectory. Due to their similarity, we count these four questions as a single template. The direction templates create a total of $16$ questions. 

\paragraph{Attribute Templates}
The second set of templates probe the models' ability to reason about relative mass, height, and circumference of common objects.
For each of these three concepts we create a set of six objects that are easily ordered by their respective attributes. A context is first presented with up to four of the six objects to prime the models with the range of possible choices. This is followed by a prompt that probes the model to select one of the objects based on the object's mass, height, or circumference. 
By inverting the superlative in the prompt (e.g., \textit{longest} $\leftrightarrow$ \textit{shortest}), we can probe the model's ability to identify both the object with the highest attribute value and the object with the lowest attribute value from the set of choices. We permute through all objects and all orders. Each of the three attributes are tested using two templates that share the same set of objects. Each template produces ${}_{6}P_{4} * 2 = 720$ questions, meaning each attribute is probed using 1440 questions. 

\paragraph{Affordance templates}
The remaining templates target an understanding of object affordances. For each affordance---stackable, rollable, graspable, breakable, slideable, and bounceable--- we collect a set of five objects with and five objects without that affordance. Again, we first provide a short context that contains each of the four possible objects. We then provide a prompt that requires the model to select the only object either with or without the affordance. We include all permutations of objects where there is exactly one correct answer. These templates produce ${}_{5}P_{1} * {}_{5}P_{3} * 4 * 2=2400$ questions for each of the six affordances.

\paragraph{Objects in PROST}
All possible values for the placeholders in our templates are shown in \cref{tab:aff_objs}. For affordances, we display objects in two groups: those with and without each affordance. For attributes, objects are sorted by increasing order, e.g., for mass, \emph{leaf} is the lightest object and \emph{microwave} is the heaviest object. Each object in PROST is selected to be single-token compatible for a wide range of vocabularies to enable easy probing of MLMs. We validate the order of our attribute objects and the group membership for our affordance objects by collecting judgments from 9  human validators. The validators obtained a 100\% agreement on the object ordering, and 94.6\% agreement on the object group membership.

\begin{table}[H]\footnotesize
\setlength{\tabcolsep}{2pt}
\centering
\begin{tabular}{l|l} \toprule
\multicolumn{2}{c}{\textbf{Attributes}} \\
\midrule
\textit{Attribute} & \textit{Objects} \\
\midrule
\textbf{mass} & leaf, coin, egg, apple, brick, microwave \\
\textbf{height} & book, microwave, table, car, house, mountain \\
\textbf{circ} & book, microwave, table, car, house, mountain \\
\midrule
\multicolumn{2}{c}{\textbf{Affordances}} \\\midrule 
\textit{Affordance} &  \textit{Objects} \\\midrule 
\textbf{stack} & books, blocks, boxes, coins, plates \\
\textbf{no stack} & balls, bottles, eggs, flowers, lamps\\ \midrule
\textbf{roll} & apple, ball, bottle, egg, can\\
\textbf{no roll} & book, box, block, mirror, microwave \\\midrule
\textbf{grasp} & balls, blocks, books, bottles, flowers\\
\textbf{no grasp} & flour, rice, salt, snow, sugar \\\midrule
\textbf{break} & bottle, egg, glass, mirror, plate\\
\textbf{no break} & ball, coin, pen, pillow, shirt \\\midrule
\textbf{slide} & ice, frost, grease, oil, soap\\
\textbf{no slide}&  carpet, concrete, grass, gravel, rubber \\\midrule
\textbf{bounce} & asphalt, brick, concrete, rubber, steel\\
\textbf{no bounce} & carpet, foam, grass, leaves, snow \\\bottomrule
\end{tabular}
\caption{Objects used in the templates.}
\label{tab:aff_objs}
\end{table}

\section{Models}
Using PROST, we probe three types of transformer-based models \citep{transformers}: decoder models, 
encoder models, 
and encoder-decoder models. 
Each model has slightly different formatting requirements, which we show in \cref{tab:processing}.  
For each model type, we probe a range of different sizes to investigate the effects of scaling. 
We use Huggingface's \citep{huggingface} pretrained models, see Table \ref{tab:modelsizes} for the full set. 

\begin{table}[th!]\centering \footnotesize
\begin{tabular}{rl|r|r}
\toprule
\multicolumn{2}{c|}{\textbf{Model}} & \multicolumn{1}{c|}{\textbf{Params (M)}} & \multicolumn{1}{c}{\textbf{Data (GB)}} \\
\midrule
GPT &         &  116.5 &              2 \\
\midrule
GPT-2 & B      &  124.4 &             40\\
& M           &  354.8 &             40\\
& L           &  774.0 &             40\\
& XL          & 1557.6 &             40\\
\midrule
BERT & B      &  110.1 &             13 \\
& L           &  336.2 &             13 \\
\midrule
RoBERTa & B   &  124.7 &            160 \\
& L           &  355.4 &            160 \\
\midrule
ALBERT V2 & B &   11.8 &            160 \\
& L           &   17.8 &            160 \\
& XL          &   59.0 &            160 \\
& XXL         &  223.2 &            160 \\
\midrule
T5 & S        &   60.5 &            170 \\
& B           &  222.9 &            170 \\
& L           &  737.7 &            170 \\
& 3B          & 2851.6 &            170 \\
\bottomrule
\end{tabular}
\caption{Summary of models evaluated on PROST. We list the amount of pretraining data as the size of the uncompressed text corpus used. }
\label{tab:modelsizes}
\end{table}

\paragraph{Decoder Models}
We analyze OpenAI's GPT-1 \citep{gpt} and GPT-2 \citep{gpt2}. 
Both are based on a transformer decoder architecture and trained on a traditional language modeling objective. We run these models over each question four times, each time with a different choice replacing the [MASK] token. Following \citet{blimp}, we select the sentence with the highest probability.

\paragraph{Encoder Models}
We analyze BERT (uncased) \citep{bert}, RoBERTa \citep{roberta}, and ALBERT \citep{albert}, which are all based on transformer encoders. 
BERT is trained on MLM and next sentence prediction and uses static masking, RoBERTa is trained on MLM with dynamic masking, and ALBERT uses whole-word $n$-gram masking. 
For probing, we filter out all but the four answer choices from the output vocabulary and select the token with the highest probability as the model's decision.

\paragraph{Encoder-decoder Models}
\label{sec:clm}
We also include results for T5 \citep{t5}. T5 is trained using a span corruption objective, in which spans of the input sequence are randomly replaced with a single mask token. During pretraining, span lengths are chosen randomly with an average length of three. To keep our results consistent with the other models, we restrict the span length to one token. We find that two of the options for sliding surfaces, namely \emph{ice} and \emph{frost}, violate our single-token constraint. To avoid any unfair comparison between answers that differ in token lengths and following previous work \citep{goldberg}, we chose to omit presenting the results for T5 on the sliding concept.

\begin{table*}[th!]\centering \footnotesize
\setlength{\tabcolsep}{2pt}
\begin{tabular}{rl|cccccccccc|c}
\toprule
\multicolumn{2}{c|}{\textbf{Model}} & \textbf{Direction} & \textbf{Mass} & \textbf{Height} & \textbf{Circum.} & \textbf{Stack} & \textbf{Roll} & \textbf{Grasp} & \textbf{Break} & \textbf{Slide} & \textbf{Bounce} & \textbf{Macro Average} \\
\midrule
GPT &  & 46.7 & 40.1 & 24.3 & 22.8 & 28.2 & 27.9 & 19.6 & 22.7 & 14.6 & 14.4 & 26.1 \\
\midrule
GPT2 &  & 43.3 & 31.4 & 22.0 & 18.8 & 26.2 & 20.3 & 17.9 & 22.5 & 16.9 & 17.0 & 23.6 \\
& M & 48.3 & 34.1 & 21.6 & 21.6 & 25.5 & 23.7 & 24.9 & 27.8 & 22.5 & 18.5 & 26.8 \\
& L & 46.7 & 33.1 & 25.4 & 27.0 & 25.5 & 26.9 & 20.6 & 21.8 & 21.3 & 15.6 & 26.4 \\
& XL & 46.7 & 34.2 & 25.8 & 26.3 & 31.1 & 36.3 & 29.4 & 26.7 & 23.7 & 20.5 & 30.1 \\
\midrule
BERT & B & 40.0 & 32.9 & 27.5 & 25.6 & 20.9 & 26.1 & 23.3 & 28.0 & 18.2 & 13.0 & 25.5 \\
& L & \textbf{70.0} & 38.8 & 19.4 & 17.5 & 21.3 & 19.2 & 26.7 & 19.5 & 15.9 & 18.6 & 26.7 \\
\midrule
RoBERTa & B & 46.7 & 36.9 & 25.8 & 23.5 & 34.5 & 19.3 & 25.4 & \textbf{45.0} & 20.9 & 11.4 & 28.9 \\
& L & 66.7 & \textbf{43.4} & 33.8 & 22.7 & 22.7 & 22.2 & 29.4 & 23.8 & 22.7 & 25.5 & 31.3 \\
\midrule
ALBERT V2 & B & 21.7 & 35.4 & 30.2 & 26.0 & 25.2 & 32.5 & 35.3 & 22.8 & 15.3 & 22.9 & 26.7 \\
& L & 41.7 & 38.2 & 31.9 & 27.5 & 23.3 & 29.7 & 34.0 & 24.5 & 23.4 & 22.1 & 29.6 \\
& XL & 46.7 & 38.7 & \textbf{42.0} & \textbf{37.4} & 30.2 & 28.2 & \textbf{37.1} & 17.8 & \textbf{25.3} & 14.3 & \textbf{31.8} \\
& XXL & 68.3 & 33.8 & 28.1 & 24.5 & 29.4 & 23.4 & 21.2 & 30.2 & 17.5 & 22.1 & 29.8 \\
\midrule
T5 & S & 20.0 & 36.5 & 29.8 & 25.2 & 25.0 & 25.9 & 25.4 & 25.0 & --- & 30.2 & 27.0$^*$ \\
& B & 40.0 & 37.0 & 32.6 & 23.8 & 25.0 & 23.4 & 25.2 & 25.6 & --- & \textbf{37.8} & 30.1$^*$ \\
& L & 46.7 & 35.7 & 30.7 & 27.6 & 31.8 & 23.0 & 34.0 & 25.2 & --- & 22.7 & 30.8$^*$ \\
& 3B & 46.7 & 39.6 & 35.6 & 29.8 & \textbf{34.7} & 31.5 & 35.6 & 33.8 & --- & 12.5 & 33.3$^*$ \\
\midrule
UnifiedQA & S & 0.0 & 34.2 & 34.8 & 30.3 & 24.4 & 29.0 & 28.8 & 27.1 & --- & 31.0 & 26.6$^*$ \\
& B & 0.0 & 17.8 & 33.3 & 22.3 & 25.5 & 34.9 & 27.9 & 36.5 & --- & 45.7 & 27.1$^*$ \\
& L & 83.3 & 17.2 & 49.5 & 47.3 & 23.5 & 28.4 & 27.5 & 43.6 & --- & 32.6 & 39.2$^*$ \\
& 3B & 63.3 & 37.8 & 55.2 & 66.9 & 31.2 & 35.2 & 24.8 & 81.4 & --- & 24.8 & 46.7$^*$ \\
\midrule
\multicolumn{2}{c|}{\textbf{Task Average}} & 46.3 & 36.5 & 28.6 & 25.2 & 27.1 & 25.9 & 27.4 & 26.0 & 19.9 & 19.9 & 28.5 \\
\bottomrule
\end{tabular}

\caption{Macro average for each concept and overall for each model on PROST. The best accruacy for general pretrained-only models is displayed in bold. Note that the task average does not include UnifiedQA.
}
\label{tab:fullresults}
\end{table*}

\paragraph{Finetuned Conditional LMs}
To better understand the limitations of text-only training, we additionally evaluate UnifiedQA \cite{unifiedqa}. UnifiedQA is a pretrained QA model, built off T5, and finetuned on SQuad 1.1, SQuaD 2.0, NarrativeQA, RACE, ARC, OpenBookQA, MCTest, and BoolQ \cite{squad1,squad2,NarrativeQA,RACE,ARC,OpenBookQA,MCTest,BoolQ}. We format all of our templates to fit their multiple-choice question answering format and use their provided scoring metrics to select the models' answers.\footnote{\href{https://github.com/allenai/unifiedqa}{https://github.com/allenai/unifiedqa}}

\section{Results}
The per model and per concept results are shown in \cref{tab:fullresults}.
For concepts with more than one template---direction, mass, height, and circumference---we average across templates to get the concept's score.

We can see that, on average, ALBERT-V2-XL performs best, with an accuracy of $31.8\%$\footnote{Note: as detailed in \cref{sec:clm}, T5 and UnifiedQA are not being evaluated on sliding. We therefore disregard their average accuracy.}, and GPT-2 performs worst, with an accuracy of $23.6\%$. We note that random guessing would yield an accuracy of $25\%$. Furthermore, every model underperforms random guessing on at least one concept. Since PROST is trivially solvable for humans, this supports our hypothesis that pretrained models are unable to perform physical reasoning anywhere close to human performance.

Comparing across all concepts, we see that direction obtains the highest average accuracy with $46.8\%$. The second best accuracy is observed for the mass attribute with $36.5\%$. 
The concepts models struggle the most with are the slideable and bounceable affordances, both with an average accuracy of $19.9\%$. 

\label{sec:position}
\begin{table}[ht!]\centering \small
\setlength{\tabcolsep}{8.5pt}
\begin{tabular}{l|cccc}\toprule
\multirow{2}{*}{\textbf{Model}} & \multicolumn{4}{c}{\textbf{Position Accuracy}} \\ 
\cmidrule(r){2-5}
& \textbf{1} & \textbf{2} & \textbf{3} & \textbf{4} \\
\midrule
GPT       & 27.0 & 24.3 &  7.6 & 38.6 \\
GPT-2     & 29.9 & 23.1 &  8.1 & 42.0 \\
BERT      & 28.4 & 24.3 &  5.7 & 38.2 \\
RoBERTa   & 39.0 & 28.7 & 11.2 & 30.0 \\
ALBERT V2 & 32.5 & 25.8 &  9.7 & 44.2 \\
T5        & 52.4 & 21.1 &  1.9 & 35.2 \\
UnifiedQA & 41.0 & 27.7 & 18.8 & 51.9 \\
\midrule
\textbf{Position Average} & 35.7 & 25.0 &  9.0 & 40.0 \\
\bottomrule
\end{tabular}
\caption{Accuracy across the correct answer's position in the context.\label{position_acc}} 
\end{table}

\begin{table*}[t]\centering \footnotesize
\setlength{\tabcolsep}{3.5pt}
\begin{tabular}{rl|ccccccccc|c}
\toprule
\multicolumn{2}{c|}{\textbf{Model}} & \textbf{Mass} & \textbf{Height} & \textbf{Circum.} & \textbf{Stack} & \textbf{Roll} & \textbf{Grasp} & \textbf{Break} & \textbf{Slide} & \textbf{Bounce} & \textbf{Macro Average} \\
\midrule
GPT &  & 9.4 & 2.2 & 14.0 & 35.9 & 43.0 & 22.7 & 13.3 & 9.8 & 10.1 & 17.8 \\
\midrule
GPT-2 &  & 19.2 & 24.3 & 1.1 & 16.1 & 5.1 & 12.1 & 31.9 & 24.9 & 15.8 & 16.7 \\
& M & 31.2 & 12.9 & 21.5 & 12.7 & 20.2 & 7.8 & 49.9 & 33.3 & 9.2 & 22.1 \\
& L & 25.8 & 24.2 & 25.4 & 5.6 & 16.7 & 18.4 & 24.6 & 28.2 & 23.7 & 21.4 \\
& XL & 43.5 & 6.7 & 1.5 & 56.1 & 51.5 & 36.3 & 31.5 & 15.8 & 32.4 & 30.6 \\
\midrule
BERT & B & 5.0 & 40.0 & 2.5 & 12.3 & 15.4 & 3.1 & 44.9 & 12.2 & 11.2 & 16.3 \\
& L & 19.2 & 21.5 & 5.8 & 1.8 & 3.4 & 4.2 & 17.2 & 9.5 & 30.4 & 12.6 \\
\midrule
RoBERTa & B & 4.7 & 4.0 & 6.5 & 55.0 & 13.8 & 27.8 & 89.6 & 21.8 & 15.8 & 26.5 \\
& L & 31.0 & 24.7 & 26.8 & 9.7 & 21.1 & 33.2 & 31.5 & 33.9 & 33.2 & 27.2 \\
\midrule
ALBERT V2 & B & 4.7 & 31.0 & 7.9 & 14.8 & 14.4 & 66.4 & \textbf{2.3} & \textbf{7.2} & 1.5 & 16.7 \\
& L & \textbf{0.6} & 11.9 & 23.6 & 30.0 & 36.8 & 52.2 & 6.9 & 28.7 & 13.2 & 22.7 \\
& XL & 9.4 & \textbf{0.7} & 8.5 & 19.5 & 8.1 & 31.0 & 10.9 & 8.2 & 20.6 & 13.0 \\
& XXL & 18.1 & 2.9 & 18.5 & 4.2 & 12.2 & 2.0 & 37.2 & 16.3 & 11.3 & 13.6 \\
\midrule
T5 & S & 8.3 & 12.9 & 13.5 & \textbf{0.0} & 3.9 & 0.8 & 3.1 & --- & 1.8 & \textbf{5.5} \\
& B & 8.7 & 26.9 & 5.8 & 0.0 & \textbf{0.2} & \textbf{0.5} & 3.1 & --- & 26.2 & 8.9 \\
& L & 5.0 & 20.3 & 1.7 & 7.5 & 22.6 & 10.5 & 37.7 & --- & \textbf{0.5} & 13.2 \\
& 3B & 16.1 & 12.2 & \textbf{0.7} & 9.2 & 8.8 & 5.1 & 34.7 & --- & 24.9 & 14.0 \\
\midrule
UnifiedQA & S & 46.9 & 5.4 & 19.7 & 2.8 & 34.5 & 31.0 & 39.0 & --- & 9.3 & 23.6 \\
& B & 19.2 & 19.2 & 4.3 & 24.6 & 30.4 & 10.4 & 4.9 & --- & 49.5 & 20.3 \\
& L & 18.5 & 28.5 & 26.2 & 18.6 & 28.1 & 41.6 & 7.9 & --- & 48.1 & 27.2 \\
& 3B & 8.2 & 46.2 & 36.1 & 6.8 & 1.8 & 0.7 & 13.4 & --- & 26.4 & 17.5 \\
\midrule
\multicolumn{2}{c|}{\textbf{Task Average}} & 15.3 & 16.4 & 10.9 & 17.1 & 17.5 & 19.7 & 27.7 & 19.2 & 16.6 & 17.6 \\

\bottomrule
\end{tabular}

\caption{Absolute difference in accuracy between a question and its superlative inverse.\label{tab:inversion}}
\end{table*}

\section{Analysis}
\paragraph{Object Order in Context}
For the concepts that use objects, all four choices are listed in each question's context. PROST contains all permutations with regards to their ordering. 
This enables us to directly look at the effect of the correct answer's position within the context on the models' accuracy. These results shown in \cref{position_acc}.

We see that models have a strong tendency to select either the first or the last item seen in the context. The largest difference is found for T5, with an accuracy of $52.4\%$ for objects at position 1 and an accuracy of only $1.9\%$ for objects at position 3. 
We note that a proper understanding of the questions, as most humans would have, would be robust to the order in which the choices are presented. This further underlines that state-of-the-art models do not perform human-like physical reasoning. 

\begin{figure}[th!]
    \centering
    \includegraphics[width=\columnwidth]{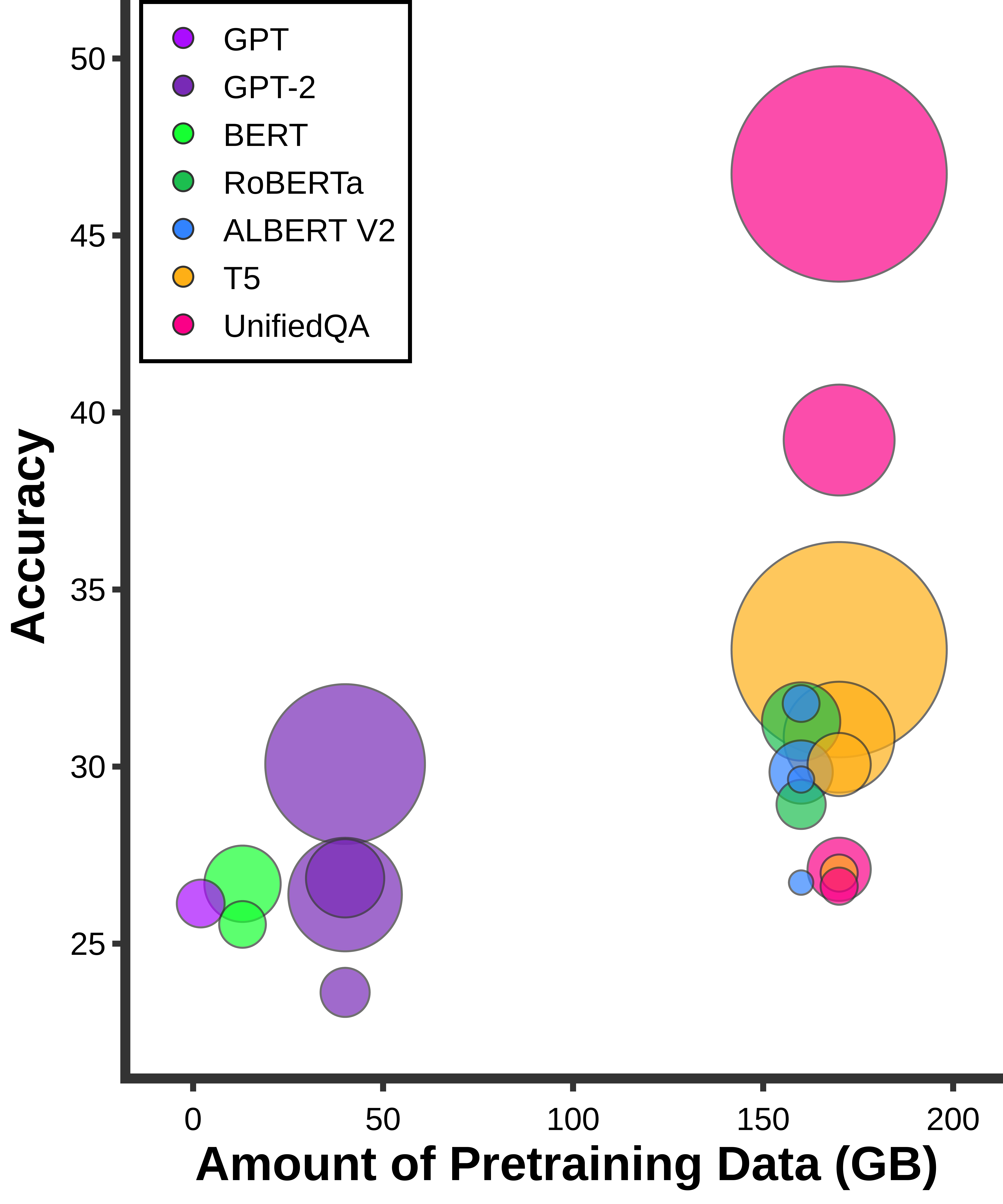}
    \caption{Scaling effect of models on accuracy. Circles size represents number of parameters.} 
    \label{fig:scaling}
\end{figure}

\paragraph{Superlative Inverses}
By inverting the superlative in a question, we are able to probe a mirrored version of the question. For example, for attributes, this would require the model to identify the lightest object instead of the heaviest object, or, for affordances, it would require the model to identify the not stackable object instead of the stackable object. We call these mirrored versions \emph{superlative inverses}.
A true understanding of the questions in PROST should be robust to this kind of inversion. However, \cref{tab:inversion} shows all models perform better on one of the two versions. Of the probed models, GPT-2 is the most unbalanced, averaging 30.6\% higher for one version over the other.

\paragraph{Data and Model Scaling}
\Cref{fig:scaling} shows each model's accuracy as a function of the number of its parameters. Unlike for many modern benchmarks, where increasing the number of parameters or training data provides significant benefits \citep{olmpics,GLUE}, PROST does not see much improvement from such scaling. We observe some improvements with T5-3B outperforming T5-small, but this 6.6\% increase requires a ~48x increase in parameters and T5-small still outperforms T5-3B on one task. Moreover, some models break this trend: ALBERT's XL version outpeforms its XXL counterpart and GPT-2 M outperforms GPT-2 L. 
While previous work has revealed the impressive scaling laws of transformer-based architectures \citep{scalinglaws}, PROST highlights the importance of relevant and informative training. As physical reasoning is not an ability that humans acquire via text, even substantially more open domain textual data is unlikely to lead to more than marginal improvements. 

\paragraph{The Limits of Text-based Training}
To our knowledge, UnifiedQA is the most qualified model to succeed on our task, having been finetuned on a significant amount of relevant text data. While this additional data does provide benefits on PROST, it still falls short, with the best performing model we tested only achieving a 46.7\% accuracy. Additionally, from \cref{position_acc,tab:inversion}, it still lacks the robustness of proper understanding. This emphasizes that models are unlikely to obtain human-like reasoning from text-based training alone. Rather, PROST motivates exposing models to concepts through multiple modalities that mirror a human's experience.

\begin{figure}[th!]
    \centering  
    \includegraphics[width=\columnwidth]{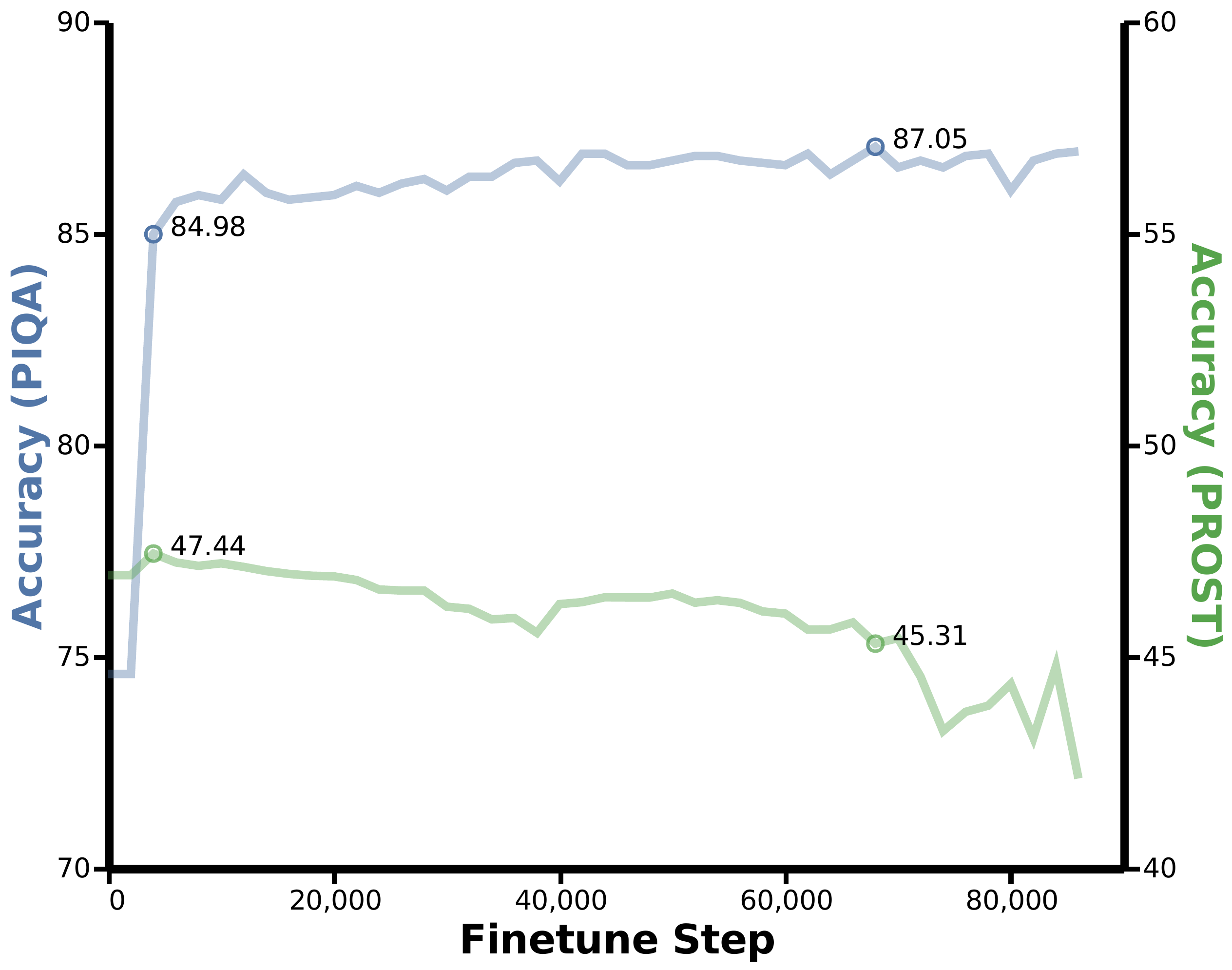}
    \caption{Analysis of the performance of UnifiedQA 3B on PROST throughout PIQA finetuning. The left and right Y axis represent Accuracy on the PIQA dev set and Macro accuracy on PROST respectively. We finetune for 100K steps, and compute metrics every 2k steps. Annotations correspond to the checkpoints with the best performance on PIQA and PROST. Note that PIQA has two answer choices, while PROST has 4.} 
    \label{fig:prostvspiqa}
\end{figure}

\paragraph{Comparing PROST and PIQA}
Due to their shared focus on text-based physical reasoning, PROST and PIQA share similarities. To test if models trained on PIQA are able to carry over any concepts to PROST, we further finetune a UnifiedQA model on PIQA and evaluate it on PROST. The results, shown in Figure \ref{fig:prostvspiqa}, indicate that training a model on PIQA is detrimental to its performance on PROST. While PIQA and PROST share a few conceptual similarities, they differ in terms of format, style, and vocabulary. We thus hypothesize that current models learn more about these surface-level differences than the conceptual similarities underpinning the questions. We further highlight two key differences between the two datasets:
\begin{itemize}
    \item PROST probes models in a zero-shot fashion, whereas PIQA provides training and test sets of identically distributed examples. This makes it possible for models on PIQA to answer successfully using spurious correlations rather than  physical reasoning.
    \item PIQA \citep{PIQA} covers an extensive range of objects and challenging physical concepts. \citet{experience} argues that experience is a prerequisite for understanding. It is hard to imagine how to expose a model to experiences ranging from egg yolk separation to making a pillow out of a garbage bag. In contrast, PROST provides a clear set of well defined concepts and objects that a model could potentially experience.
\end{itemize}

\section{Discussion}
Our experiments show that all the models we analysed fail to demonstrate a robust understanding of physical reasoning. Beyond performing poorly across every concept, they are easily influenced by changing the order of the objects in a question's context and by superlative inverses. Moreover, our analysis indicates that these issues are not likely to be solved simply by increasing the amount of model parameters or training data. All this evidence supports \citet{climbing}'s and \citet{experience}'s theory that experience is a prerequisite of understanding. 

A number of other reasoning benchmarks have been solved to some extent by a large finetuned model. UnifiedQA (11B parameters), based on T5 \cite{t5}, achieved 81.4\% on ARC \cite{ARC}; and UNICORN
\footnote{\href{https://leaderboard.allenai.org/hellaswag/submissions/public}{leaderboard.allenai.org/hellaswag/submissions/public}}
(11B parameters), also based on T5, achieved a 93.9\% accuracy on hellaSWAG \cite{hellaswag}. While all these models are larger and are trained on more data, our results force us to ask the question whether they perform well because these additional parameters and data have imbued the models with an ability to reason, or if they succeed by finding subtle unintended correlations in the data. This forces us to look more closely at how models succeed, and not just the accuracy they achieve. Tools like CheckList \citep{checklist} can aid in this endeavor by demonstrating how robust models are to changes in the distribution of the data. 

\paragraph{How to Use this Probe}

PROST is intended to help analyze any model that can be deployed in a text-only setting. However, we maintain that multi-modal data is necessary to experience the concepts in PROST, and that these experiences are likely a crucial step in succeeding on this dataset. One way that multi-modal models could prepare for this type of text-only evaluation is through multi-task training, where one of the tasks is only conditioned on text. Such an approach has already been considered: \citet{gpt3} propose an extension to their CLIP model 
which is trained on multiple modalities in a multi-task fashion.
Because of the templated nature of PROST, its exact format can be adapted to match specific styles of language training, as we do for T5 and UnifiedQA. 

PROST’s language-only approach is motivated by two reasons. First, we believe that true multi-modal models should be able to function on any subset of their modalities. We note that humans can easily interact with text-only inputs (e.g., a text message) while still learning from and interacting with other modalities. Second, it enables the comparison of models trained using different modalities or domains. For example, we believe comparing how language understanding modules evolve when trained on vision-and-language navigation compared to visual question answering would provide invaluable insights.

\paragraph{Limitations } We caution that achieving a high accuracy on PROST does not necessarily guarantee that a model is able of physical reasoning. It is likely easy to succeed on this benchmark if one were to intentionally train models on similar enough sentences or a subset of PROST itself. We 
hope that the community will use this dataset in the intended way: in a zero-shot setting to probe models which have been trained on data not specifically collected to succeed on PROST.

\section{Conclusion}
We present a probing dataset called PROST, which is designed to test a model's ability to reason about the physical world. Our experiments show that current state-of-the-art pretrained models lack the ability to reason about physical interactions. Further, all models struggle when the order of options is changed and when questions are inverted, both things that would not confuse humans. Lastly, our analysis shows that these issues are unlikely to be solved by simply scaling models. 
Our results highlight the need to look beyond text-based pretraining and to provide models with the necessary experiences for human-like understanding of the physical world.

\section*{Acknowledgments}
We would like to thank the members of CU Boulder's NALA and HIRO Groups for their feedback on this work.

\bibliographystyle{acl_natbib}
\bibliography{anthology,acl2021}

\end{document}